\documentclass[11pt]{article}

\usepackage[final]{acl}

\usepackage{amsmath}
\usepackage{amsfonts}
\usepackage{xcolor}
\usepackage[most]{tcolorbox}
\usepackage{fontawesome5}

\usepackage{times}
\usepackage{latexsym}

\usepackage{booktabs}
\usepackage{multirow}

\usepackage[table]{xcolor}  

\usepackage{arydshln}     

\usepackage[T1]{fontenc}

\usepackage[utf8]{inputenc}

\usepackage{microtype}

\usepackage{inconsolata}

\usepackage{graphicx}

\usepackage{amsmath}

\usepackage{algorithm}
\usepackage{algpseudocode}

\title{Stable Adaptive Thinking via Advantage Shaping and Length-Aware Gradient Regulation}

\author{
  \textbf{Zihang Xu}%
  {\hypersetup{linkcolor=black}\thanks{Equal contribution.}},
  \textbf{Haozhi Xie}%
  {\hypersetup{linkcolor=black}\footnotemark[1]},
  \textbf{Ziqi Miao}%
  {\hypersetup{linkcolor=black}\footnotemark[1]},
  \textbf{Wuxuan Gong},
  \textbf{Chen Qian},
  \textbf{Lijun Li}
\\
\\
  Beihang University,
  Shanghai AI Laboratory,\\
  Beijing University of Posts and Telecommunications,
  Renmin University of China
}

\begin{document}
\maketitle
\begin{abstract}

Large reasoning models (LRMs) achieve strong performance through extended reasoning traces, but they often exhibit overthinking behavior for low-complexity queries. Existing efforts to mitigate this issue are fundamentally limited by unstable accuracy–efficiency trade-offs and poor robustness to heterogeneous reasoning behaviors. To address these challenges, we propose a two-stage framework for stable adaptive thinking in LRMs. The framework first applies Hybrid Fine-Tuning to expose the model to both thinking and no-thinking behaviors, establishing well-conditioned initialization. It then performs adaptive reinforcement learning with Correctness-Preserving Advantage Shaping (CPAS) to avoid suppressing correct long-chain reasoning, and Length-Aware Gradient Regulation (LAGR) to stabilize optimization under severe reasoning-length heterogeneity. Extensive experiments on Qwen2.5-1.5B and 7B show consistent improvements over strong baselines, achieving up to +3.7/+3.6 accuracy points while reducing generated tokens by 40.6\%/43.9\%. 
Further analyses across varying problem difficulties and out-of-distribution tasks confirm the robustness and generalization of our approach.

\end{abstract}

\section{Introduction}\label{sec:Intro}

Recent advancements in large reasoning models (LRMs), such as OpenAI-o1~\cite{openai-o1} and DeepSeek-R1~\cite{guo2025deepseek}, have demonstrated remarkable progress in solving complex tasks. By explicitly generating extended reasoning traces, these models exhibit strong reasoning and generalization capabilities across a wide range of domains~\cite{LRM1}. Although long-form reasoning benefits challenging problems, existing LRMs lack difficulty-aware control ability and thus often produce unnecessarily long reasoning chains even for simple queries~\cite{survey_3, survey_4}. This overthinking issue significantly increases token usage and inference cost, posing a critical bottleneck for the practical efficiency of LRMs~\cite{overthink}.

\begin{figure}
    \centering
    \includegraphics[width=1.0\linewidth]{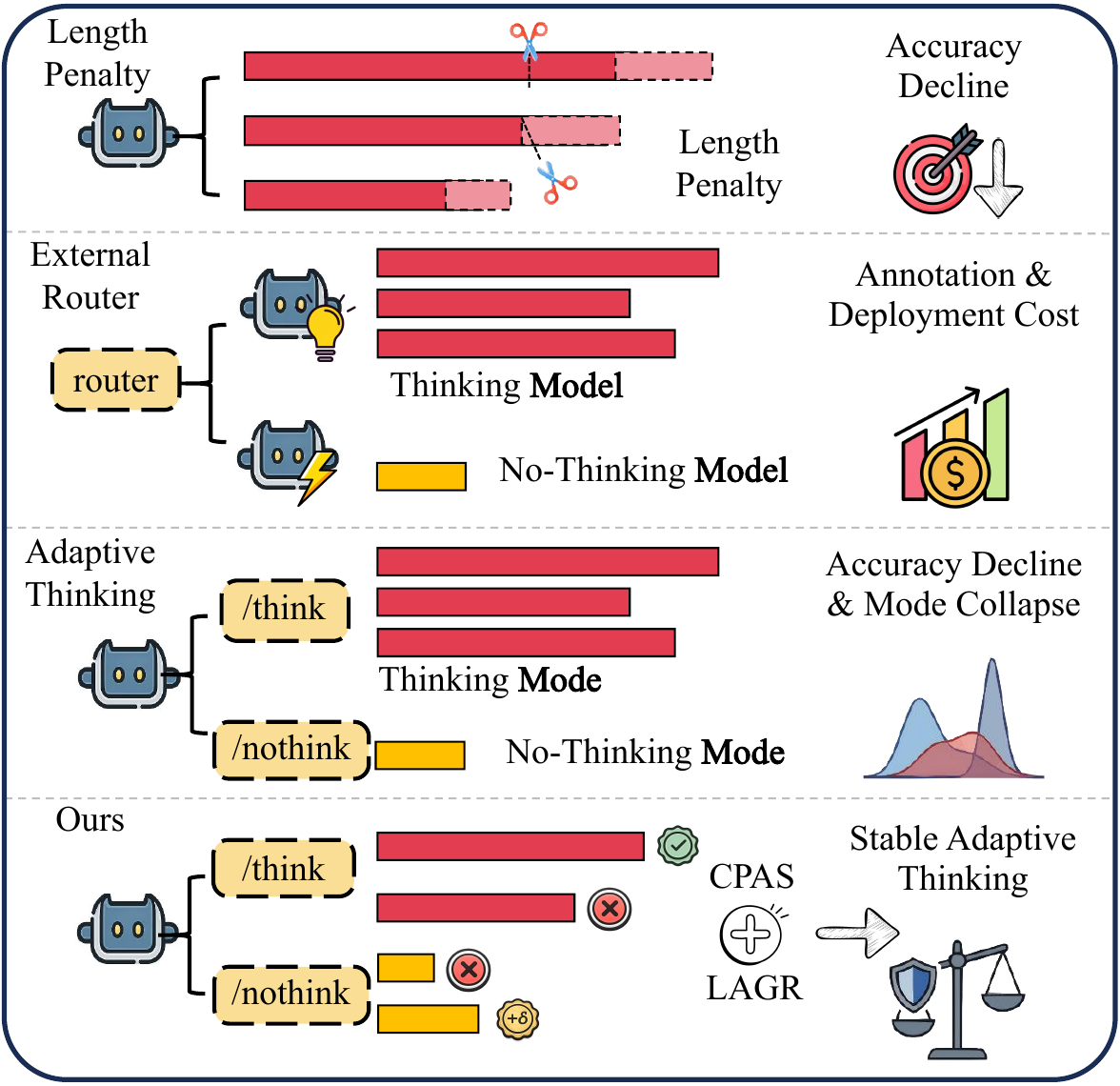}
    \caption{Illustration of efficient reasoning methods.}
    \label{fig:Intro}
\end{figure}

To mitigate the efficiency issues arising from overthinking, recent studies have investigated several efficient reasoning strategies~\cite{survey_1, survey_2}, as illustrated in Figure~\ref{fig:Intro}.
Length-based penalties have been introduced to discourage excessively long reasoning chains~\cite{O1-prune, L1, TokenSkip}; however, 
these approaches operate within the thinking mode, and 
the additional constraints in these approaches may limit the exploration space of LRMs. Routing-based methods reduce inference cost by selecting between reasoning and non-reasoning models~\cite{hu2024routerbench, RouteLLM, GPT5}, but they typically depend on external annotations for supervision and the maintenance of multiple models, thereby increasing both training and deployment cost. More recently, reinforcement-learning-based adaptive thinking methods aim to improve efficiency by encouraging shorter reasoning chains. ThinkLess~\cite{Thinkless} achieves this via a decoupled GRPO formulation, whereas AdapThink~\cite{Adaptthink} employs a PPO-based objective to promote short-chain reasoning. Despite their effectiveness, these methods often face difficulties in achieving a stable balance between reasoning accuracy and efficiency during training.

These observations highlight two key challenges in adaptive thinking.
First, there is an inherent trade-off between efficiency and accuracy. While adaptive thinking aims to reduce computational costs, an excessive bias toward brevity may constrain the model’s capacity for deep reasoning on complex tasks, ultimately degrading performance.
Second, severe length heterogeneity during training poses additional difficulties.  Within a single rollout group, reasoning chains may range from a few hundred tokens to tens of thousands of tokens, resulting in uneven optimization granularity and imbalanced gradient contributions, which may compromise training stability.

To address these challenges, we propose Stable Adaptive Thinking via Advantage Shaping and Length-Aware Gradient Regulation, a two-stage framework for stable and effective adaptive reasoning in large reasoning models. The proposed approach enables the model to adjust its reasoning depth according to input difficulty, improving efficiency on simple queries while preserving accuracy on complex ones. In the first stage, \textbf{Hybrid Fine-Tuning} is employed to establish well-conditioned initialization by exposing the model to both short and long reasoning behaviors. In the second stage, adaptive thinking is optimized through reinforcement learning with two key mechanisms. \textbf{Correctness-Preserving Advantage Shaping}, in contrast to prior approaches that directly reward short chains, introduces an advantage shaping strategy that avoids suppressing correct long-chain reasoning trajectories when encouraging efficiency, thereby preserving the model’s exploration capacity. \textbf{Length-Aware Gradient Regulation} addresses optimization instability arising from the severe reasoning length heterogeneity by regulating gradient allocation across different responses, while ensuring that control signals remain effective under long and heterogeneous reasoning sequences, leading to more stable training dynamics. These components provide a principled solution for stable adaptive thinking under efficiency–accuracy trade-offs. In general, the contributions of this paper can be summarized as follows:

\begin{itemize}

  \item We address key optimization challenges in adaptive thinking. Specifically, we focus on balancing computational efficiency with reasoning accuracy, and on mitigating training instability arising from severe length heterogeneity across different reasoning modes during reinforcement learning optimization.
  
  \item We propose a two-stage framework for stable adaptive thinking. The framework integrates hybrid fine-tuning with adaptive reinforcement learning, while employing correctness-preserving advantage shaping and length-aware gradient regulation for stable training.
  
  \item Extensive experiments on the Qwen2.5 series across multiple model scales and mathematical benchmarks demonstrate that our approach consistently outperforms existing baselines, achieving up to +3.7/+3.6 accuracy improvements while reducing the number of generated tokens by 40.6\%/43.9\%. 
  Further analyses across varying problem difficulties and diverse tasks confirm the robustness and generalization of our approach.
\end{itemize}

\section{Preliminaries}\label{sec:Preliminary}

\begin{figure*}
    \centering
    \includegraphics[width=1.0\linewidth]{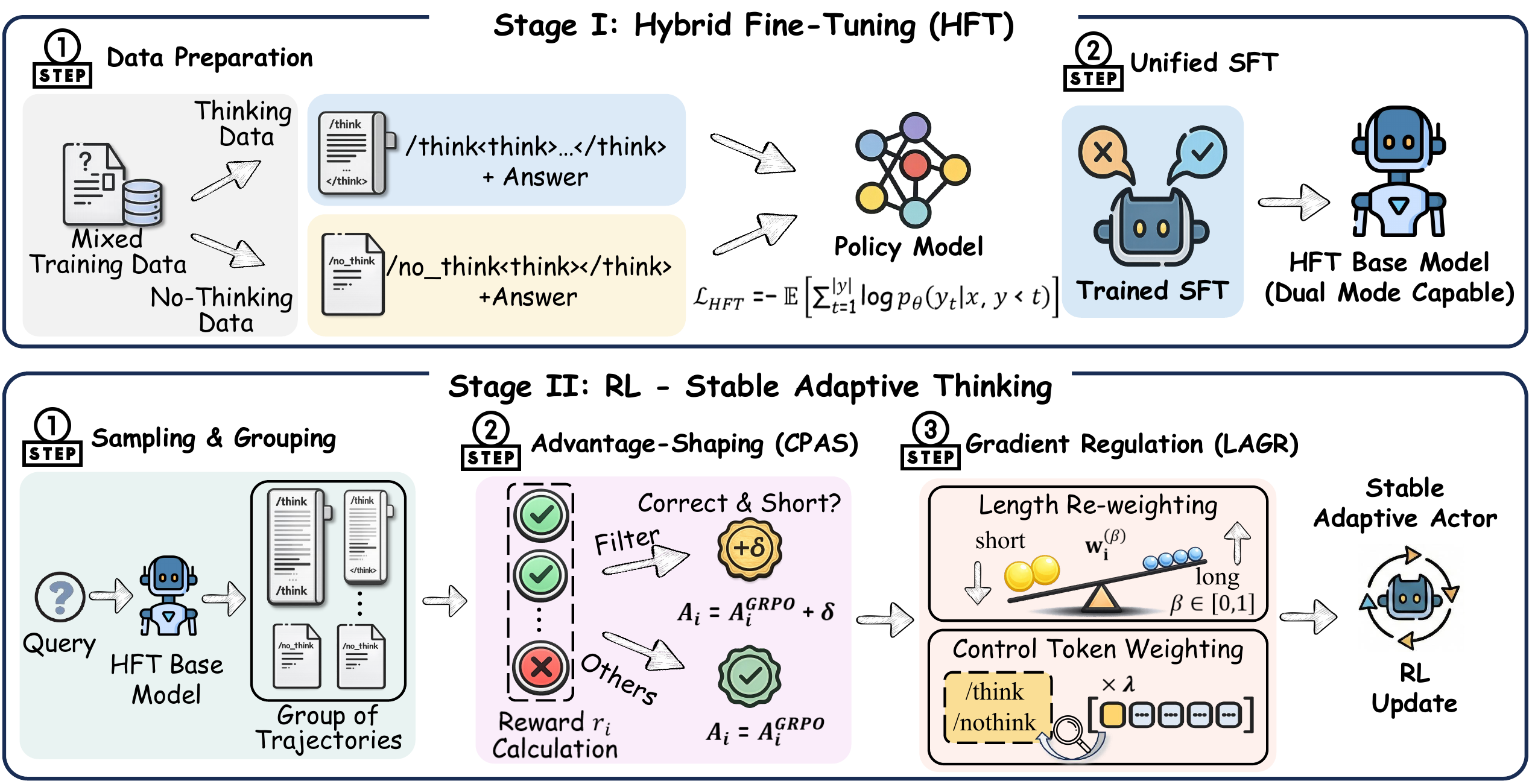}
    \caption{Overview of our two-stage training pipeline. Stage 1 performs Hybrid Fine-Tuning (HFT) on paired thinking and no-thinking formats to initialize a unified policy. Stage 2 applies GRPO-style reinforcement learning with correctness-preserving advantage shaping (CPAS) and length-aware gradient regulation (LAGR) to stabilize optimization under extreme length heterogeneity and to learn when to think.}
    \label{fig:framework}
\end{figure*}

\subsection{Adaptive Thinking}
Specifically, given an input problem $x$, the model produces an output $y$ consisting of a chain-of-thought trajectory and a final answer: $y=\{S,a\}$. We introduce a control token to specify the thinking mode $m \in \mathcal{M}=\{\texttt{/think},\texttt{/no\_think}\}$: when the control token is \texttt{/think}, the model generates a non-empty chain-of-thought trajectory $S$ in the thinking mode; when the control token is \texttt{/no\_think}, the chain-of-thought is empty $S=\emptyset$ and the model directly produces the final answer $a$.

We implement adaptive thinking with a single policy model. Given input $x\sim\mathcal{D}$, the model first generates a control token $m\in\mathcal{M}$, and then generates the remaining output conditioned on $m$. Thus, the same parameters $\theta$ define both mode selection $\pi_{\theta}$ (the distribution of the first control token) and response generation $P_\theta(y\mid x,m)$.

\subsection{Problem Statement}

The goal of adaptive thinking is to reduce unnecessary generation while preserving correctness. We define a utility function that trades off accuracy and efficiency. For an output $y=\{S, a\}$, we use an accuracy reward $R_{\text{acc}}(y\mid x)\in\{0,1\}$ indicating whether the final answer $a$ is correct (verified by an automatic checker), and a length reward $R_{\text{len}}(y\mid x)$ that favors shorter outputs. The overall utility is
\begin{align*}
    U(x,y)=R_{\text{acc}}(y\mid x)+\gamma\,R_{\text{len}}(y\mid x)
\end{align*}
where $\gamma>0$ controls the efficiency–accuracy trade-off. Our objective is to learn a policy that selects $m$ and generates $y$ to maximize expected utility:
\begin{align*}
    \max_{\theta}\ \mathbb{E}_{x\sim\mathcal{D}}\Big[\ \mathbb{E}_{m\sim\pi_\theta(\cdot\mid x)}\ \mathbb{E}_{y\sim P_\theta(\cdot\mid x,m)}\big[U(x,y)\big]\ \Big].
\end{align*}

\section{Methodologies}\label{sec:method}

\subsection{Overall Framework}
We propose a novel framework for \textbf{\emph{stable adaptive thinking}}, which enables a single model to dynamically regulate its thinking depth according to problem difficulty. The objective is to avoid unnecessary overthinking on simple queries while preserving strong deliberative capability on challenging problems. Starting from a pretrained base language model, our framework adopts a two-stage training paradigm that comprises hybrid fine-tuning for capability warm-up and reinforcement learning for stable adaptive thinking. An overview of the training pipeline is illustrated in Figure~\ref{fig:framework}.

\paragraph{Stage I: Hybrid Fine-Tuning.}  
In the first stage, we perform hybrid fine-tuning on the base model using large-scale mathematical reasoning data with heterogeneous supervision, including long CoT solutions and short direct answers distilled from strong teacher models. This stage injects both thinking and no-thinking capabilities into a unified model and provides a stable warm-up for subsequent reinforcement learning.

\paragraph{Stage II: Stable Adaptive thinking with Reinforcement Learning.}  
In the second stage, the model is further optimized with a modified GRPO algorithm, where it explores both thinking and no-thinking trajectories and learns \emph{when to think}. To stabilize training under severe length heterogeneity and prevent mode collapse, we introduce correctness-preserving advantage shaping and length-aware gradient regulation, which together facilitate effective exploration and robust training.

\subsection{Hybrid Fine-Tuning for Warm-Up}
Hybrid Fine-Tuning (HFT) is designed to provide a stable cold-start initialization for adaptive reinforcement learning by exposing the base model to both thinking and no-thinking generation patterns in a unified and format-consistent manner. Unlike the subsequent stage, HFT does not aim to learn an adaptive decision policy; instead, it focuses on expanding the model’s solution space to support heterogeneous response behaviors.

\paragraph{Hybrid-Formatted Data Construction.}
The hybrid fine-tuning dataset is constructed from large-scale open-source mathematical corpora. Each training instance is annotated with an explicit control token that specifies the desired response mode. 

For \emph{thinking-mode} examples, responses are distilled from DeepSeek-R1-0528. Each response begins with the prefix \texttt{/think} and contains an explicit thinking block of the form
\texttt{<think>\{\textit{thinking process}\}</think>},
where the content enclosed by the \texttt{<think>} tags corresponds to the intermediate thinking steps produced by the teacher model. The final answer is generated after the thinking block following a unified output structure.

For \emph{no-thinking-mode} examples, responses are distilled from DeepSeek-V3-0324. These responses begin with the prefix \texttt{/no\_think} and include an empty thinking block of the form
\texttt{<think></think>},
which explicitly indicates that no intermediate thinking is required and that a direct answer should be produced.

\paragraph{Optimization Objective.}
During HFT, the model is optimized using standard supervised learning objectives over both response types, without imposing any explicit preference between thinking and no-thinking outputs. Given the hybrid-formatted dataset $\mathcal{D}_{\text{HFT}} = \{(x_i, y_i)\}_{i=1}^{N}$, the model is optimized to maximize the conditional likelihood of the target response given the input:
{
\begin{align*}
    \mathcal{L}_{\text{HFT}} = - \mathbb{E}_{(x,y)\sim\mathcal{D}_{\text{HFT}}} \left[ \sum_{t=1}^{|y|} \log p_\theta(y_t \mid x, y_{<t}) \right].   
\end{align*}
}

\paragraph{Role of HFT.}
From a methodological perspective, HFT defines the feasible action space for subsequent reinforcement learning by ensuring that both thinking and no-thinking trajectories are readily accessible to the policy. This hybrid-formatted warm-up substantially reduces exploration difficulty in the reinforcement learning stage and provides a well-conditioned initialization.

\subsection{Stable Adaptive thinking with Reinforcement Learning}
After HFT, the model is further trained to acquire \emph{adaptive thinking} capability through reinforcement learning. While HFT equips the model with both thinking-mode and no-thinking-mode generation behaviors, it does not enable the model to autonomously decide which mode to adopt for a given query. This decision-making ability is learned in the reinforcement learning stage.

Specifically, we build upon GRPO and train the model to explore both thinking and no-thinking trajectories for each query. In this setting, the policy learns to decide \emph{whether to engage in explicit thinking} by selecting between thinking-mode and no-thinking-mode responses. However, directly applying vanilla GRPO often leads to training instability and mode collapse. We identify two key challenges that arise from GRPO under heterogeneous thinking trajectories, and introduce corresponding mechanisms to address them.

\paragraph{Correctness-Preserving Advantage Shaping.}

Existing adaptive thinking approaches often discourage overthinking by assigning higher rewards to shorter thinking trajectories. Under GRPO-style group-wise optimization, advantages are computed via intra-group normalization across multiple trajectories sampled for the same input. Let $\{r_i\}_{i=1}^{G}$ denote the rewards of a rollout group $G$, and the corresponding advantages be 
$A_i = \frac{r_i - \mathbb{Mean}(r)}{\mathbb{Std}(r)}$.
When shorter trajectories systematically receive higher rewards~\cite{L1, Thinkless}, this normalization induces a relative comparison bias: longer but correct thinking trajectories may satisfy $0 < r_i < \mathbb{Mean}(r)$ and thus receive $A_i < 0$. Consequently, valid thinking behaviors are actively suppressed during optimization, limiting exploration and potentially leading to mode collapse toward no-thinking responses.

To alleviate this issue, we introduce \emph{Correctness-Preserving Advantage Shaping} (CPAS), which reshapes advantages for correct responses based on their thinking mode.  Let $A_i^{\text{GRPO}}$ denote the advantage computed by GRPO for response $o_i$, and let $r_i\in \{0,1\}$ denote the reward. The final advantage is defined as
{
\begin{align*}
A_i =
\begin{cases}
A_i^{\text{GRPO}} + \delta,
& \text{if } r_i = 1 \text{ and } o_i \text{ is short-chain}, \\
A_i^{\text{GRPO}}, & \text{otherwise}.
\end{cases}
\end{align*}
}

where, $\delta > 0$ is a small reward coefficient that provides additional encouragement for correct short-chain responses, while leaving correct long-chain responses unaffected.

\paragraph{Role of CPAS.}
By reshaping advantages for correct responses, CPAS ensures that correct thinking trajectories receive non-negative advantage signals.
This prevents correct long-chain responses from being inadvertently penalized, thereby preserving the model’s exploration capacity.
Consequently, CPAS contributes to more stable training dynamics and enhances the effectiveness of adaptive thinking.

\paragraph{Length-Aware Gradient Regulation.}
Under GRPO, policy optimization is performed by sampling a group $G$ of responses
$\{o_i\}_{i=1}^{G}$ for each query $q$ and aggregating token-level policy gradients
within each response.
The response-level gradient contribution of $o_i$ can be written as
\begin{equation*}
g_i^{\text{GRPO}}
=
\frac{1}{|o_i|}
\sum_{t=1}^{|o_i|}
\hat{A}_{i,t}\,
\nabla_\theta
\log \pi_\theta(o_{i,t}\mid q,o_{i,<t}),
\label{eq:grpo_response_grad}
\end{equation*}
where $|o_i|$ denotes the response length and $\hat{A}_{i,t}$ is the token-level
advantage.

In adaptive thinking, rollout groups often exhibit severe length
heterogeneity, ranging from short no-thinking responses to long thinking
trajectories.
The $1/|o_i|$ normalization systematically
attenuates gradient magnitudes for long responses, increases gradient variance,
and biases optimization toward short trajectories, which may lead to premature
collapse of thinking behaviors.

To mitigate this imbalance, we propose \emph{Length-Aware Gradient Regulation}, which explicitly reweights response-level gradient contributions according to response length. Specifically, LAGR employs a length-dependent weighting scheme:
\begin{align*}
g_i^{\text{LAGR}}
=
w_i^{(\beta)}
\sum_{t=1}^{|o_i|}
\hat{A}_{i,t}\,
\nabla_\theta
\log \pi_\theta(o_{i,t}\mid q,o_{i,<t}),
\nonumber \\
w_i^{(\beta)}
=
\frac{1}{M}
\cdot
\frac{|o_i|^{-\beta}}{\sum_{j=1}^{G}|o_j|^{-\beta}},
\qquad
\beta \in [0,1].
\end{align*}

\noindent where $M$ is a constant scaling factor used to control the overall magnitude of the gradient. When $\beta=1$, LAGR recovers the GRPO-like length-sensitive behavior that favors
short responses, while $\beta=0$ assigns uniform weights across responses,
thereby alleviating the suppression of long ones.
By adjusting $\beta$, LAGR provides a mechanism to balance optimization stability and thinking flexibility.

While LAGR mitigates length-induced imbalance at the response level, adaptive
thinking introduces an additional challenge at the token level.
In practice, the decision of whether to engage in thinking is primarily encoded by the prefix control token.
However, when responses are long, the gradient signal of this control token can
be diluted by the large number of subsequent tokens. To preserve this adaptive control signal, we further apply an explicit boosting to the control token.
Let $t=0$ denote the control token.
The final response-level gradient under LAGR is given by
\begin{equation*}
\begin{array}{@{}c@{}}  
\nabla_\theta \mathcal{J}_{\text{LAGR}}(\theta) = \nabla_\theta \sum_{i=1}^{G} w_i^{(\beta)} \Bigg[ \lambda\, \min\!\Big( r_{i,0}(\theta)\,\hat{A}_{i,0},\; \\
\quad \mathrm{clip}\big(r_{i,0}(\theta),1-\epsilon,1+\epsilon\big)\,\hat{A}_{i,0} \Big) + \sum_{t=1}^{|o_i|} \\
\quad \min\!\Big( r_{i,t}(\theta)\,\hat{A}_{i,t},\; \mathrm{clip}\big(r_{i,t}(\theta),1-\epsilon,1+\epsilon\big)\,\hat{A}_{i,t} \Big) \Bigg].
\end{array}
\end{equation*}
\noindent where $\lambda$ is a control-token boosting factor that compensates for gradient
dilution in long responses.

\paragraph{Role of LAGR.}
LAGR rebalances response-level policy gradients under length heterogeneity, mitigating GRPO’s bias toward short trajectories.
By preserving effective gradient signals for long thinking chains and boosting the control token, LAGR prevents decision-signal dilution and supports stable adaptive thinking.

\section{Experiments}\label{sec:Experiment}

\subsection{Experimental Settings}

\begin{table*}[t]
\caption{Overall performance on MATH-500, AIME-2024, and AIME-2025 for Qwen2.5 series models. $Len$ is the average number of generated tokens per example. $Ratio_{NT}$ denotes the fraction of examples solved in NoThinking mode. $\Delta Acc$ and $\Delta Len$ are computed relative to the always-Thinking baseline of the same base model.}
\label{tab:main_results}
\centering
\small
\setlength{\tabcolsep}{5pt}
\renewcommand{\arraystretch}{1.15}
\newcommand{\fade}[1]{\textcolor{black!45}{#1}}

\begin{tabular}{ll ccc ccc ccc cc}
\toprule
\multicolumn{2}{c}{\multirow{2}{*}{\textbf{Method}}} &
\multicolumn{3}{c}{\textbf{MATH-500}} &
\multicolumn{3}{c}{\textbf{AIME-2024}} &
\multicolumn{3}{c}{\textbf{AIME-2025}} &
\multicolumn{2}{c}{\textbf{Average}} \\
\cmidrule(lr){3-5}
\cmidrule(lr){6-8}
\cmidrule(lr){9-11}
\cmidrule(lr){12-13}
\multicolumn{2}{c}{} &
\textbf{Acc} & \textbf{Len} & \textbf{\(\text{Ratio}_{NT}\)} &
\textbf{Acc} & \textbf{Len} & \textbf{\(\text{Ratio}_{NT}\)} &
\textbf{Acc} & \textbf{Len} & \textbf{\(\text{Ratio}_{NT}\)} &
$\Delta$\textbf{Acc} & $\Delta$\textbf{Len} \\
\midrule

\multicolumn{13}{c}{\textbf{Qwen-2.5-1.5B}} \\
\cmidrule(lr){1-13}

& \fade{Thinking}
& \fade{80.4} & \fade{10441} & \fade{0.0\%}
& \fade{25.7} & \fade{35829} & \fade{0.0\%}
& \fade{20.6} & \fade{36977} & \fade{0.0\%}
& \fade{0.0} & \fade{+0.0\%} \\

& \fade{No-Thinking}
& \fade{60.4} & \fade{2320} & \fade{100.0\%}
& \fade{9.1} & \fade{10751} & \fade{100.0\%}
& \fade{7.5} & \fade{11150} & \fade{100.0\%}
& \fade{-16.6} & \fade{-70.9\%} \\

& \fade{HFT}
& \fade{72.3} & \fade{7106} & \fade{42.4\%}
& \fade{16.8} & \fade{23673} & \fade{54.2\%}
& \fade{14.5} & \fade{25376} & \fade{46.2\%}
& \fade{-7.4} & \fade{-32.5\%} \\

& O1-Pruner
& 77.1 & 7043 & 0.0\%
& 20.3 & 33417 & 0.0\%
& 16.7 & 32185 & 0.0\%
& -4.2 & -12.7\% \\

& RouteLLM
& 76.7 & \textbf{3862} & 81.6\%
& 19.8 & 28305 & 30.8\%
& 17.4 & 29973 & 27.8\%
& -4.3 & -25.4\% \\

& AdaptThink
& \underline{81.1} & 4341 & 69.4\%
& \underline{25.6} & \textbf{22819} & 4.2\%
& \underline{22.7} & \underline{24187} & 3.3\%
& +0.9 & -38.3\% \\

& Thinkless
& 78.5 & 4473 & 71.3\%
& 24.5 & 25294 & 2.9\%
& 19.3 & 26351 & 2.5\%
& -1.5 & -32.6\% \\

\rowcolor{gray!12}
& \textbf{Ours}
& \textbf{84.6} & \underline{4176} & 69.8\%
& \textbf{28.1} & \underline{23258} & 3.7\%
& \textbf{25.2} & \textbf{22062} & 5.4\%
& \textbf{+3.7} & \textbf{-40.6\%} \\

\midrule

\multicolumn{13}{c}{\textbf{Qwen-2.5-7B}} \\
\cmidrule(lr){1-13}

& \fade{Thinking}
& \fade{90.6} & \fade{8309} & \fade{0.0\%}
& \fade{65.6} & \fade{29936} & \fade{0.0\%}
& \fade{59.3} & \fade{35217} & \fade{0.0\%}
& \fade{0.0} & \fade{+0.0\%} \\

& \fade{No-Thinking}
& \fade{75.1} & \fade{1890} & \fade{100.0\%}
& \fade{25.5} & \fade{10464} & \fade{100.0\%}
& \fade{20.2} & \fade{9921} & \fade{100.0\%}
& \fade{-31.6} & \fade{-70.1\%} \\

& \fade{HFT}
& \fade{86.3} & \fade{5619} & \fade{45.2\%}
& \fade{50.9} & \fade{23484} & \fade{42.5\%}
& \fade{44.2} & \fade{25017} & \fade{40.4\%}
& \fade{-11.3} & \fade{-27.3\%} \\

& O1-Pruner
& 87.9 & 5567 & 0.0\%
& 62.4 & 25977 & 0.0\%
& 57.5 & 32451 & 0.0\%
& -2.6 & -12.9\% \\

& RouteLLM
& 88.2 & \textbf{3121} & 81.8\%
& 55.3 & 24011 & 31.2\%
& 51.7 & 28387 & 27.1\%
& -6.8 & -24.4\% \\

& AdaptThink
& \underline{91.1} & 3462 & 77.8\%
& \underline{65.2} & \underline{18123} & 12.1\%
& \underline{59.1} & 22544 & 10.4\%
& 0.0 & -39.9\% \\

& Thinkless
& 89.8 & \underline{3302} & 78.6\%
& 64.0 & 18172 & 13.8\%
& 60.1 & \underline{21463} & 12.5\%
& -0.5 & -41.6\% \\

\rowcolor{gray!12}
& \textbf{Ours}
& \textbf{92.4} & 3403 & 77.6\%
& \textbf{69.2} & \textbf{17762} & 12.5\%
& \textbf{64.6} & \textbf{20060} & 9.6\%
& \textbf{+3.6} & \textbf{-43.9\%} \\

\bottomrule
\end{tabular}
\end{table*}

\paragraph{Models.} 
We adopt Qwen-2.5-1.5B-Base and Qwen-2.5-7B-Base as the initial policy models~\cite{qwen2025qwen25technicalreport}. Both are extensively pre-trained and serve as the foundation for jointly learning thinking and no-thinking behaviors.

\paragraph{Dataset and Metrics.} 
For HFT, we construct a large-scale training corpus from open-source mathematics datasets that require advanced numerical reasoning and multi-step logical inference, including OpenR1-Math-220k~\cite{faceopen}, DeepMath-103K~\cite{he2025deepmath}, and NuminaMath~\cite{jia2024numinamath}, among others. Paired supervision is obtained by distilling thinking-mode outputs from DeepSeek-R1-0528~\cite{guo2025deepseek} and no-thinking outputs from DeepSeek-V3-0324~\cite{deepseekv3}. In the RL stage, we primarily adopt the DeepScaleR~\cite{luo2025deepscaler} dataset. For evaluation, we focus on standard mathematical reasoning benchmarks, including MATH-500~\cite{lightman2023let}, AIME-2024~\cite{AM24}, and AIME-2025~\cite{AM25}, and additionally employ GPQA~\cite{rein2024gpqa} to assess out-of-distribution generalization. We report accuracy, average token length, and no-thinking ratio to jointly measure model performance and inference efficiency.

\begin{figure*}[ht]
    \centering
    \includegraphics[width=1\linewidth]{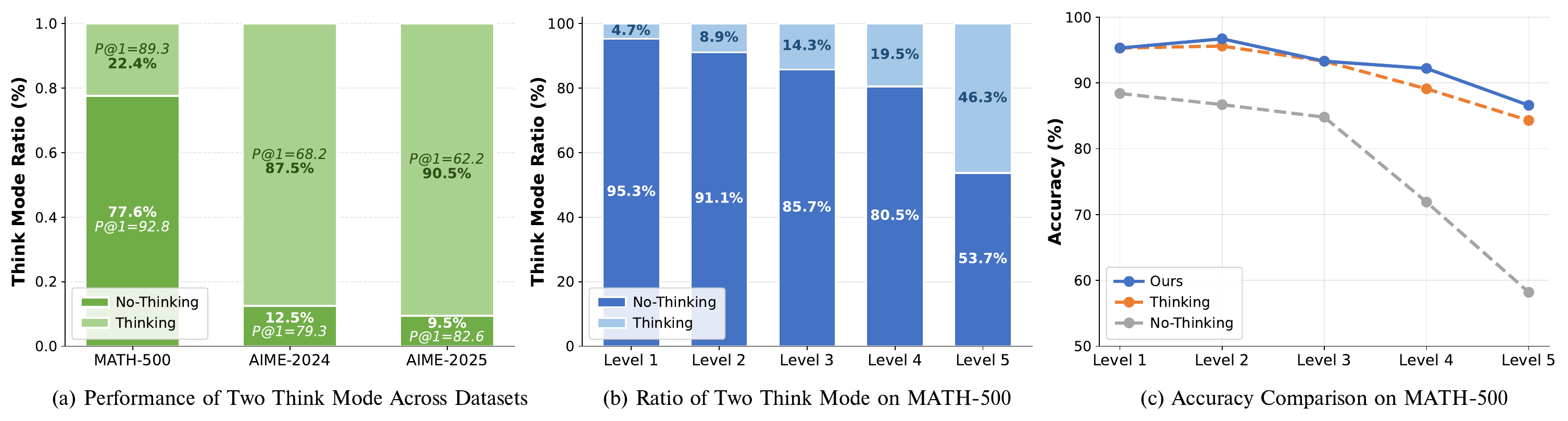}
    \caption{Difficulty-aware mode selection and performance. (a) Performance of Two Think Mode on MATH-500, AIME-2024, and AIME-2025. (b) Mode ratio across MATH-500 difficulty levels. (c) Accuracy across difficulty levels, comparing our adaptive policy with always-Thinking and always-No-Thinking baselines.}
    \label{fig:adapt_think}
\end{figure*}

\begin{figure*}
    \centering
    \includegraphics[width=1.02\linewidth]{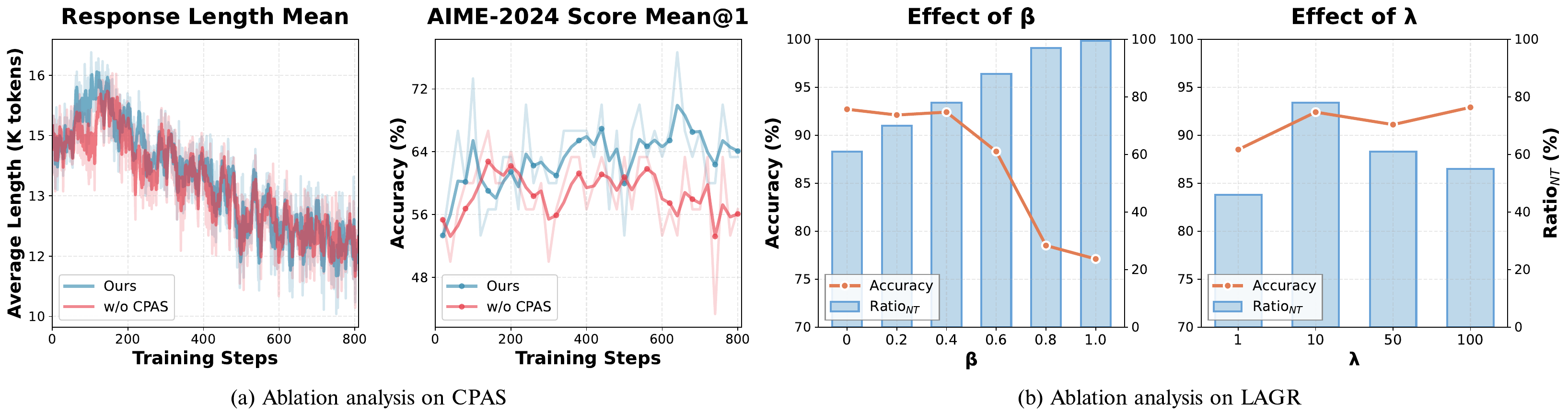}
    \caption{Ablation and sensitivity analysis. (a) Training dynamics with/without CPAS: mean response length (left) and AIME-2024 accuracy (right). (b) Effect of the LAGR length-weight parameter $\beta$ (left) and the control-token boost factor $\lambda$ (right) on accuracy and the no-thinking ratio.}
    \label{fig:ablation}
\end{figure*}

\paragraph{Baselines.} 
To ensure a comprehensive comparison, we evaluate our method on both 1.5B and 7B model scales and consider four categories of baselines. 
(1) Base Variants: thinking-only, no-thinking-only, and HFT hybrid models derived from the same base architectures.  (2) Length-Penalization Methods: O1-Pruner~\cite{O1-prune} encourages shorter thinking processes based on a thinking-only model under accuracy constraints. (3) Routing Methods: RouteLLM~\cite{RouteLLM} trains a router to dynamically choose between long and short CoT using separate thinking and no-thinking models. (4) Adaptive Thinking Methods: ThinkLess~\cite{Thinkless} learns a hybrid thinking policy via a decoupled GRPO strategy, while AdaptThink~\cite{Adaptthink} promotes concise thinking using PPO-based optimization. Both are initialized from the HFT models.
This ensures a fair comparison across methods with consistent base architectures and appropriate training settings.

\paragraph{Implementation Details.}
All experiments are conducted on 16 NVIDIA H100 GPUs. In the Hybrid Fine-Tuning stage, we set the maximum context length to 16K tokens, with overlong samples truncated accordingly. The models are trained for 3 epochs using the LLaMA-Factory~\cite{llamafactory} framework. Training is conducted with the AdamW optimizer, employing a 10\% linear warmup strategy followed by cosine learning rate decay, with the maximum learning rate set to 1e-4.
In the reinforcement learning stage, the context length is extended to 24K tokens. Experiments are conducted using the VeRL~\cite{Verl} framework, with the policy model optimized by AdamW at a constant learning rate of 1e-6. We use a batch size of 128 with a micro-batch size of 64, and sample 8 responses per query during rollout. For verification, we adopt the Math-Verify module to provide correctness-based feedback. More details can be found in the Appendix.

\subsection{Main Results}

In this section, we conduct a comprehensive evaluation of different models on three mathematical reasoning benchmarks. As shown in Table~\ref{tab:main_results}, our method achieves consistent gains at both the 1.5B and 7B scales, improving performance by 3.7 and 3.6 points, respectively, while reducing token costs by 40.6\% and 43.9\%, demonstrating both accuracy and efficiency improvements.
Compared to length-penalty methods, our approach leverages an explicit no-thinking mode that does not constrain exploration, thereby achieving lower token consumption and improved overall performance. Compared to external routing methods, our approach requires no additional annotated data or deployment overhead, and can more accurately select appropriate thinking modes based on the model’s capability. Finally, relative to existing adaptive thinking methods, the proposed correctness-preserving advantage shaping and length-aware gradient regulation achieve significantly better performance under comparable generation lengths, enabling stable and effective adaptive thinking.

\subsection{Analysis of Adaptive Thinking}

We further analyze model behavior across problems of varying difficulty to evaluate the effectiveness of adaptive thinking. As shown in Figure~\ref{fig:adapt_think}(a), on the relatively easy MATH-500 benchmark, the proposed method selects the no-thinking mode for up to 77.6\% of the queries, whereas on the more challenging AIME-2024 and AIME-2025 benchmarks, it activates the thinking mode more frequently. Notably, performance under the no-thinking mode consistently surpasses that under the thinking mode, indicating that the model learns to adaptively select the more appropriate strategy according to problem difficulty. Figures~\ref{fig:adapt_think}(b) and~\ref{fig:adapt_think}(c) further illustrate the model’s thinking behavior across different difficulty levels within MATH-500. As the difficulty increases, the model progressively increases its reliance on the thinking mode. Compared with the original model that exclusively uses either the thinking or no-thinking mode, our method consistently achieves superior performance across all difficulty levels. These results demonstrate that the proposed approach attains a more favorable balance between reasoning efficiency and accuracy.

\subsection{Ablation and Sensitivity Analysis}

\paragraph{Effect of CPAS.}
Figure~\ref{fig:ablation}(a) illustrates the training dynamics with and without CPAS. Compared with the baseline, our method exhibits a faster increase in response length at early stages and reaches a higher peak, indicating earlier and deeper exploration. The length then gradually decreases and converges to a similar level, achieving stable adaptive thinking behavior. Evaluation on the AIME benchmarks further shows that our method consistently outperforms the baseline, demonstrating a better balance between accuracy and efficiency, which confirms the effectiveness of CPAS.

\paragraph{Effect of LAGR.}
We analyze the impact of LAGR under different parameters. As shown in Figure~\ref{fig:ablation}(b) left, values of $\beta$ closer to 1 encourage no-thinking chains, whereas values closer to 0 promote the thinking exploration. An intermediate setting of $\beta = 0.4$ provides the best trade-off between accuracy and efficiency. Figure~\ref{fig:ablation}(b) right examines the effect of control-token weighting. When the weight is set to 1, the ratio between thinking and no-thinking modes remains largely unchanged, indicating that the control signal is diluted. Increasing the weight to 10 enables stable adaptive thinking, while larger weights induce overly aggressive updates, prematurely assigning solvable samples to the thinking mode and reducing efficiency.

\subsection{Generalization Analysis}
We evaluate out-of-distribution (OOD) generalization on the GPQA benchmark, which consists of graduate-level multiple-choice questions spanning diverse scientific domains and differs from the training data in both question format and content.
As shown in Table~\ref{tab:ood-gpqa}, our method achieves the highest accuracy while reducing the average response length by 51.0\%. The model activates the no-thinking mode for 31.3\% of the queries, yielding substantial efficiency gains without compromising performance. These results demonstrate that our approach generalizes effectively to OOD settings, enabling robust and efficient adaptive thinking.

\begin{table}[t]
  \centering
  \setlength{\tabcolsep}{4.5pt}
  \caption{OOD Performance on GPQA Dataset.}
  \label{tab:ood-gpqa}
  \begin{tabular}{lcccc}
    \toprule
    \multirow{2}{*}{Method} &
    \multicolumn{4}{c}{GPQA} \\
    \cmidrule(lr){2-5}
    & Acc & Len & Ratio$_{NT}$ & $\Delta$Len \\
    \midrule
    Thinking   & 47.5 & 24135 & 0.0\%    & --      \\
    No-Thinking & 35.3 & 8031  & 100.0\%  & -66.7\% \\
    HFT        & 41.8 & 16849  & 44.1\%  & -30.2\% \\
    \textbf{Ours} & 50.4 & 11826  & 31.3\% & -51.0\% \\
    \bottomrule
  \end{tabular}
\end{table}

\section{Related Work}\label{sec:Related Work}

\subsection{Large Reasoning Models}

Large reasoning models~\cite{openai-o1, guo2025deepseek} have recently attracted substantial attention due to their superior performance.
Enabled by scalable reinforcement learning~\cite{he2025deepmath, DAPO, GSPO}, these models can generate long reasoning chains using only outcome-based and format rewards, leading to improved performance in complex reasoning tasks. To further generalize in various scenarios, previous work has explored reinforcement learning with verifiable rewards~\cite{RLPR} and introduced explicit supervision over the reasoning process~\cite{PRM}. However, most existing approaches overlook the efficiency issues caused by long redundant reasoning chains, which increases the inference cost. Creating efficient reasoning without sacrificing performance remains an open challenge.

\subsection{Efficient Reasoning for LRMs}

Efficient reasoning for LRMs aims to reduce overthinking while preserving accuracy. One line of work focuses on compressing or regularizing reasoning traces via prompting, supervised tuning with variable-length CoT, or length-regularized RL \cite{overthink, COTValve, L1, O1-prune}. These methods typically assume reasoning is always necessary and may restrict the exploration. Another line studies hybrid thinking via routing, allocating compute by switching between thinking and no-thinking models based on task difficulty \cite{SelfRoute, Thinkswitcher, GPT5}, but it often requires external supervision and deployment costs. More recently, RL-based adaptive thinking learns when to think using explicit control tokens or rewards for no-thinking behaviors \cite{Thinkless, HGPO, Adaptthink, AdaCOT, DRSAF}. However, achieving stable optimization under the efficiency–accuracy trade-off remains challenging, motivating our stability-oriented approach.

\section{Conclusion}\label{sec:Conclusion}

In this work, we address the training instability caused by heterogeneous reasoning lengths in existing adaptive thinking methods, as well as the difficulty of jointly optimizing reasoning accuracy and efficiency. We propose a novel two-stage framework for stable adaptive thinking. In Stage I, we adopt HFT to endow the model with both thinking and no-thinking modes. In Stage II, CPAS preserves the model’s ability to explore long reasoning chains, while LAGR stabilizes reinforcement learning training. Extensive experiments demonstrate that our approach achieves a better balance between performance and efficiency, underscoring the potential of our proposed advantage-based model optimization strategy for reinforcement learning.

\section*{Limitations}\label{sec:Limitation}
In this section, we discuss several limitations of this work.
First, due to computational constraints, our empirical evaluation is limited to models at the 1.5B and 7B scales. While consistent improvements across these settings demonstrate the effectiveness of the proposed approach, experiments on larger models would be valuable for more thoroughly assessing its scalability.
Second, our training primarily relies on mathematical reasoning datasets, which offer reliable and verifiable reward signals. Although evaluations on GPQA indicate promising out-of-distribution generalization, extending training to more diverse domains with trustworthy verification mechanisms may further enhance robustness and applicability.

\bibliography{custom}

\clearpage

\newpage
\setcounter{page}{1} 
\renewcommand\thesection{\Alph{section}}
\setcounter{section}{0}

\section*{Appendix}\label{sec:Appendix}

\section{Experimental Details}

\subsection{Data Construction}
\label{app:data_construction}

To initialize both thinking and no-thinking behaviors in a unified policy, we construct a \emph{hybrid-formatted training corpus} that explicitly contains paired supervision for the two reasoning modes.

\paragraph{Data Sources.}
The HFT dataset is built from large-scale public mathematical corpora, including OpenR1-Math, DeepMath-103K, and NuminaMath, which cover diverse problem types requiring numerical computation and multi-step reasoning.

\paragraph{Hybrid Supervision Construction.}
For each problem, we construct paired supervision corresponding to two reasoning modes:
\begin{itemize}
    \item \textbf{Thinking mode.} Long-form trajectories are distilled from DeepSeek-R1-0528, prefixed with the control token \texttt{/think} and containing an explicit reasoning block \texttt{<think>...\ </think>} followed by the final answer.
    \item \textbf{No-thinking mode.} Short direct-answer responses are distilled from DeepSeek-V3-0324, prefixed with \texttt{/no\_think} and including an empty reasoning block \texttt{<think></think>}.
\end{itemize}

The two types of supervision are balanced with a \emph{1:1 ratio} to avoid introducing bias toward either reasoning mode during supervised warm-up.

\paragraph{Verification.}
All constructed responses are verified using the Math-Verify module, and samples that fail verification are discarded. This step ensures correctness and reliability of both thinking and no-thinking supervision.

Overall, after verification and filtering, we obtain a hybrid corpus consisting of approximately 600K thinking samples and 600K no-thinking samples, providing clean, balanced supervision for initializing adaptive thinking policies.

\subsection{Baselines}
\label{app:baselines}
\begin{table*}[t!]
\caption{Appendix Results on Diverse Model Sizes and Architectures.}
\label{tab:appendix_results}
\centering
\small
\setlength{\tabcolsep}{5pt}
\renewcommand{\arraystretch}{1.15}

\begin{tabular}{ll ccc ccc ccc}
\toprule
\multicolumn{2}{c}{\multirow{2}{*}{Method}} &
\multicolumn{3}{c}{MATH-500} &
\multicolumn{3}{c}{AIME-2024} &
\multicolumn{3}{c}{AIME-2025} \\
\cmidrule(lr){3-5}
\cmidrule(lr){6-8}
\cmidrule(lr){9-11}
\multicolumn{2}{c}{} &
Acc & Len & $\text{Ratio}_{NT}$ &
Acc & Len & $\text{Ratio}_{NT}$ &
Acc & Len & $\text{Ratio}_{NT}$ \\
\midrule

\multicolumn{11}{c}{Qwen-2.5-1.5B} \\
\cmidrule(lr){1-11}

& Thinking
& 80.4 & 10441 & 0.0\%
& 25.7 & 35829 & 0.0\%
& 20.6 & 36977 & 0.0\% \\

& No-Thinking
& 60.4 & 2320 & 100.0\%
& 9.1 & 10751 & 100.0\%
& 7.5 & 11150 & 100.0\% \\

& HFT
& 72.3 & 7106 & 42.4\%
& 16.8 & 23673 & 54.2\%
& 14.5 & 25376 & 46.2\% \\

& Ours
& 84.6 & 4176 & 69.8\%
& 28.1 & 23258 & 3.7\%
& 25.2 & 22062 & 5.4\% \\

\midrule

\multicolumn{11}{c}{Qwen-2.5-7B} \\
\cmidrule(lr){1-11}

& Thinking
& 90.6 & 8309 & 0.0\%
& 65.6 & 29936 & 0.0\%
& 59.3 & 35217 & 0.0\% \\

& No-Thinking
& 75.1 & 1890 & 100.0\%
& 25.5 & 10464 & 100.0\%
& 20.2 & 9921 & 100.0\% \\

& HFT
& 86.3 & 5619 & 45.2\%
& 50.9 & 23484 & 42.5\%
& 44.2 & 25017 & 40.4\% \\

& Ours
& 92.4 & 3403 & 77.6\%
& 69.2 & 17762 & 12.5\%
& 64.6 & 20060 & 9.6\% \\

\midrule

\multicolumn{11}{c}{Llama3.1-8B} \\
\cmidrule(lr){1-11}

& Thinking
& 89.8 & 9046 & 100.0\%
& 63.3 & 31294 & 100.0\%
& 55.7 & 34801 & 100.0\% \\

& No-Thinking
& 72.1 & 2423 & 0.0\%
& 20.6 & 9112 & 0.0\%
& 17.5 & 10905 & 0.0\% \\

& HFT
& 82.2 & 5619 & 48.8\%
& 44.7 & 22388 & 46.3\%
& 37.1 & 22938 & 49.2\% \\

& Ours
& 90.8 & 3618 & 75.5\%
& 66.7 & 19089 & 10.1\%
& 60.4 & 21084 & 8.7\% \\

\bottomrule
\end{tabular}
\end{table*}

To comprehensively evaluate the effectiveness of the proposed approach, we compare against a diverse set of baselines that represent different strategies for controlling reasoning length and inference efficiency. All baselines are evaluated under the same experimental settings and verification protocols unless otherwise specified.

\paragraph{Base Variants.}
We first consider several variants derived from the same base language model to isolate the effect of reasoning behaviors:
\begin{itemize}
    \item \textbf{Thinking-only.} The model always operates in the thinking mode, generating explicit chain-of-thought reasoning for every input. This variant provides the deepest level of reasoning but incurs substantial token overhead.
    \item \textbf{No-thinking-only.} The model always generates direct answers without intermediate reasoning. While this variant is computationally efficient, it typically suffers from degraded performance on complex problems.
    \item \textbf{HFT.} The hybrid fine-tuned model trained with both thinking and no-thinking supervision, as described in Appendix~\ref{app:data_construction}. This model possesses both capabilities but does not autonomously decide which mode to use.
\end{itemize}

\paragraph{Length-Penalization Methods.}
We include methods that discourage excessively long reasoning chains by explicitly penalizing output length:
\begin{itemize}
    \item \textbf{O1-Pruner}~\cite{O1-prune}. This approach introduces length-aware constraints during fine-tuning to prune redundant reasoning steps while maintaining correctness. It operates exclusively within the thinking mode and does not support explicit no-thinking behavior.
\end{itemize}

\paragraph{Routing-Based Methods.}
Routing approaches aim to reduce inference cost by dynamically selecting between different reasoning strategies:
\begin{itemize}
    \item \textbf{RouteLLM}~\cite{RouteLLM}. This method trains an external router to select between a thinking model and a no-thinking model at inference time. While effective in reducing token usage, it relies on additional supervision for router training and requires maintaining multiple models during deployment.
\end{itemize}

\paragraph{Adaptive Thinking Methods.}
We further compare against recent reinforcement-learning-based adaptive thinking approaches:
\begin{itemize}
    \item \textbf{ThinkLess}~\cite{Thinkless}. This method adopts a decoupled GRPO formulation to encourage shorter chains, learning a hybrid policy over thinking and no-thinking behaviors.
    \item \textbf{AdaptThink}~\cite{Adaptthink}. This approach employs a PPO-based objective that explicitly promotes concise reasoning by increasing the likelihood of short chains.
\end{itemize}

Both ThinkLess and AdaptThink share the same HFT initialization as our method, ensuring fair comparison and controlled base model capacity.


\paragraph{Evaluation Protocol.}
All methods are evaluated using the same datasets, verification module (Math-Verify\footnote{https://github.com/huggingface/Math-Verify}), and decoding settings. We report accuracy, average generation length, and the no-thinking ratio to jointly measure reasoning performance and efficiency. For MATH-500, all reported results are averaged over 4 independent runs. For AIME-2024 and AIME-2025, results are averaged over 32 independent runs to reduce variance.

\subsection{Implementation Details}
\label{app:impl_details}

All experiments are conducted on 16 NVIDIA H100 GPUs. We implement the Hybrid Fine-Tuning (HFT) stage using LLaMA-Factory\footnote{https://github.com/hiyouga/LlamaFactory} and the reinforcement learning stage using VeRL\footnote{
https://github.com/volcengine/verl}.

\paragraph{HFT.}
In the HFT stage, the maximum context length is set to 16K tokens, with overlong samples truncated. Models are trained for 3 epochs with the AdamW optimizer, using a 10\% linear warmup followed by cosine learning rate decay, and a peak learning rate of $1\times 10^{-4}$.

\paragraph{Reinforcement Learning.}
In the RL stage, the context length is extended to 24K tokens. The policy model is optimized with AdamW using a constant learning rate of $1\times 10^{-6}$. We use a batch size of 128 with a micro-batch size of 64, and sample $G{=}8$ responses per query during rollout.

\section{Additional Experimental Results}
\label{app:add_results}

To further assess the generality of our approach, we conduct additional experiments beyond the Qwen2.5 series. Specifically, we apply the same training pipeline to LLaMA 3.1-8B Base, which adopts a different architecture and pretraining recipe.
All training and optimization procedures strictly follow the HFT and RL settings described in Appendix \ref{app:impl_details}, with only minimal hyperparameter adjustments for model size. No architecture-specific modifications are introduced.

As shown in Table~\ref{tab:appendix_results}, our method consistently improves reasoning accuracy while substantially reducing the average generation length under the LLaMA architecture. These results demonstrate that the proposed correctness-preserving advantage shaping and length-aware gradient regulation are not specific to a particular backbone, but instead generalize well across different model families.

\section{Case Study}

\begin{figure*}
\centering
\begin{tcolorbox}[size=title,opacityfill=0.1]
\noindent
\faUser\ \textbf{Question (MATH500)}: Convert the point $(0,3)$ in rectangular coordinates to polar coordinates. Enter your answer in the form $(r,\theta),$ where $r > 0$ and $0 \le \theta < 2 \pi.$ Format your response as follows: ``The correct answer is boxed\{insert answer here\}.'' \\

\faCogs\ \textbf{Thinking}:

\texttt{\textbf{<think>}}I need to convert the point $(0,3)$ from rectangular coordinates to polar coordinates. Polar coordinates are given as $(r,\theta)$, where $r$ is the distance from the origin to the point, and $\theta$ is the angle from the positive $x$-axis. First, recall the formulas for converting from rectangular to polar coordinates. The formulas are: $r = \sqrt{x^2 + y^2}$ and $\theta = \tan^{-1}(y/x)$. [...](\textbf{omit}) which is not the same point. So, only $\theta = \pi/2$ is correct. I could use $\theta = \pi/2 + 2k\pi$, but within $0$ to $2\pi$, only $\pi/2$. $2\pi$ would be $5\pi/2$, which is greater than $2\pi$, so no. So, I think $(3,\pi/2)$ is the answer. The problem says ``enter your answer in the form $(r,\theta)$'', and ``boxed''. So, I need to write it as $(3,\frac{\pi}{2})$ in the box. The instruction says: ``output the final answer within \texttt{\textbackslash boxed\{\}}'' and ``enter your answer as $(r,\theta)$'', so, probably $(3,\ \frac{\pi}{2})$. But let me make sure the format. Sometimes they write it as a pair. I think that's fine. Is there any other representation? For example, could $r$ be negative? But the problem specifies $r>0$, so no. Or $\theta$ could be different, but no. Another thing: when $x=0$ and $y>0$, it's unambiguously $\pi/2$. If it were $(0,-3)$, it would be $3\pi/2$ or $-\pi/2$, but since it's positive, it's fine. So, I think that's it. The point is $(0,3)$, which is on the $y$-axis, so $\theta=90$ degrees, which is $\pi/2$ radians. Yes. So, final answer.
\texttt{\textbf{</think>}}
Final Answer:
To convert the point $(0, 3)$ from rectangular coordinates to polar coordinates $(r, \theta)$ with $r > 0$ and $0 \le \theta < 2\pi$, calculate $r = \sqrt{x^2 + y^2} = \sqrt{0^2 + 3^2} = 3$. Since $x = 0$ and $y > 0$, the point lies on the positive $y$-axis, so $\theta = \pi/2$, and $\pi/2 \in [0, 2\pi)$. Thus, the polar coordinates are $(3, \pi/2)$, i.e., the correct answer is $\boxed{\left(3,\ \dfrac{\pi}{2}\right)}$. \textbf{(3272 tokens)}
\
\\

{\color[HTML]{2EA121}\faCogs\ \textbf{Ours}}: \\
\texttt{\textbf{/nothink <think></think>}} To convert the rectangular coordinates $((0, 3))$ to polar coordinates $(r, \theta)$, follow these steps: 1.\ Calculate $r$ (the distance from the origin to the point): The formula for $r$ is: $ r = \sqrt{x^2 + y^2} $ Substituting $x = 0$ and $y = 3$: $ r = \sqrt{0^2 + 3^2} = \sqrt{9} = 3 $ 2.\ Determine $\theta$ (the angle with the positive $x$-axis): Since the point $((0, 3))$ lies on the positive $y$-axis, the angle $\theta$ is: $ \theta = \frac{\pi}{2} $ This is because the positive $y$-axis corresponds to $90^\circ$ or $\left(\frac{\pi}{2}\right)$ radians in the range $0 \leq \theta < 2\pi$. 3.\ Combine the results: The polar coordinates are: $ (3, \frac{\pi}{2}) $ Final Answer: $ \boxed{\left(3, \dfrac{\pi}{2}\right)} $ \textbf{(713 tokens)}

\end{tcolorbox}
\caption{For an example from MATH-500, the Thinking baseline generates a long chain-of-thought with redundant intermediate steps. In contrast, our method chooses \textit{NoThinking} and directly produces a concise final solution, using 713 tokens in total.}
\label{case:math500_polar}
\end{figure*}

\begin{figure*}
\centering
\begin{tcolorbox}[size=title,opacityfill=0.1]
\noindent
\faUser\ \textbf{Question (AIME)}: Find the largest possible real part of $\left(75+117i\right)z+\dfrac{96+144i}{z}$ where $z$ is a complex number with $|z|=4$. \\

\faCogs\ \textbf{Thinking}:

\texttt{\textbf{<think>}} I need to find the largest possible real part of $(75 + 117i)z + \frac{96 + 144i}{z}$ where $z$ is a complex number with magnitude 4. Since $|z| = 4$, I can write $z = 4e^{i\theta}$ for some real $\theta$, or equivalently, $z = 4(\cos \theta + i \sin \theta)$. Let me denote the expression as $w = (75 + 117i)z + \frac{96 + 144i}{z}$. I need the real part of $w$, and maximize it. Since $z$ is on the circle of radius 4, I could set $z = x + iy$ with $x^2 + y^2 = 16$, but that might get messy with the algebra. Notice that the expression involves both $z$ and $1/z$, and since $|z| = 4$, $|1/z| = 1/4$. Also, the coefficients are complex numbers, so I need to handle that. Let me simplify the coefficients. \textbf{[...](omit)} So real part is Re( (r + i s)/4 * (cos $\theta$ - i sin $\theta$) ) = (1/4) Re( (r + i s)(cos $\theta$ - i sin $\theta$) ) = (1/4) [ r cos $\theta$ - r i sin $\theta$ + i s cos $\theta$ - $i^2$ s sin $\theta$ ] = (1/4) [ r cos $\theta$ + s sin $\theta$ + i ( -r sin $\theta$ + s cos $\theta$ ) ] So real part is (1/4)(r cos $\theta$ + s sin $\theta$) Similarly, from first term: 4 [ (p + i q)(cos $\theta$ + i sin $\theta$) ] = 4 [ p cos $\theta$ - q sin $\theta$ + i (p sin $\theta$ + q cos $\theta$) ], so real part 4(p cos $\theta$ - q sin $\theta$) Thus total real part Re(w) = $4(p \cos \theta - q \sin \theta) + \frac{1}{4} (r \cos \theta + s \sin \theta)$ Now plug in p=75, q=117, r=96, s=144. So Re(w) = $4(75 \cos \theta - 117 \sin \theta) + \frac{1}{4} (96 \cos \theta + 144 \sin \theta)$ Compute: 4*75=300, 4*(-117)= -468, so 300 $\cos \theta$ - 468 $\sin \theta$ Then $\frac{1}{4} *96 = 24$, $\frac{1}{4}*144=36$, so +24 $\cos \theta$ + 36 $\sin \theta$ Thus total: (300 + 24) $\cos \theta$ + (-468 + 36) $\sin \theta$ = 324 $\cos \theta$ - 432 $\sin \theta$, same as before. So no mistake. Thus maximum is $\sqrt{324^2 + (-432)^2} = \sqrt{104976 + 186624} = \sqrt{291600} = 540.$ So the largest possible real part is 540. Therefore, the answer is 540. But the problem says "real part", and I got it, so I think that's it. \texttt{\textbf{</think>}} To find the largest possible real part of $(75 + 117i)z + \frac{96 + 144i}{z}$ where $z$ is a complex number with $|z| = 4$, express $z$ in polar form as $z = 4e^{i\theta}$ for some real $\theta$. This gives $z = 4(\cos \theta + i \sin \theta)$. Define the expression as $w = (75 + 117i)z + \frac{96 + 144i}{z}$. Substitute $z$: $w = (75 + 117i) \cdot 4(\cos \theta + i \sin \theta) + \frac{96 + 144i}{4(\cos \theta + i \sin \theta)}.$ Simplify the second term: $\frac{96 + 144i}{4(\cos \theta + i \sin \theta)} = \frac{96 + 144i}{4} \cdot \frac{1}{\cos \theta + i \sin \theta} = (24 + 36i) \cdot \frac{\cos \theta - i \sin \theta}{1},$ since $\frac{1}{\cos \theta + i \sin \theta} = \cos \theta - i \sin \theta$ for $|z| = 4$. \textbf{[...](omit)}This maximum is achieved for some $\theta$, such as when $\cos \theta = \frac{3}{5}$ and $\sin \theta = -\frac{4}{5}$, giving $\operatorname{Re}(w) = 540$. Since the function is continuous and the circle $|z| = 4$ is compact, the maximum value is attained. Thus, the largest possible real part is 540. \textbf{(22383 tokens)}
\
\\

{\color[HTML]{2EA121}\faCogs\ \textbf{Ours}}: \\
\texttt{\textbf{/nothink <think></think>}} Alright, I have this problem: Find the largest possible real part of $(75+117i)z+\frac{96+144i}{z}$ where $z$ is a complex number with $|z|=4$. Understanding the Problem First, I need to understand what's being asked. We have a complex expression in terms of $z$, and $z$ is a complex number with magnitude 4. The goal is to find the maximum real part of this expression. Step 1: Represent $z$ in a Convenient Form Since $z$ is a complex number with $|z| = 4$, it can be written in polar form as: $z = 4 e^{i\theta}$ where $\theta$ is the argument of $z$. But before jumping into polar form, maybe it's easier to think of $z$ in terms of its real and imaginary parts: $z = x + iy$ where $x, y$ are real numbers, and $x^2 + y^2 = 16$ because $|z| = 4$. Step 2: Rewrite the Expression The expression is:  \textbf{[...](omit)}  $x = \frac{3}{5} \times 4 = \frac{12}{5}$ $y = \frac{-4}{5} \times 4 = \frac{-16}{5}$ Now, plug these into $(81x - 108y)$: $81 \times \frac{12}{5} - 108 \times \frac{-16}{5} = \frac{972}{5} + \frac{1728}{5} = \frac{972 + 1728}{5} = \frac{2700}{5} = 540$ This matches our previous result. Conclusion The largest possible real part of the given expression is $540$. \textbf{(2977 tokens)}

\end{tcolorbox}
\caption{For an example from AIME, the Thinking baseline generates a long chain-of-thought with redundant intermediate steps. In contrast, our method chooses \textit{NoThinking} and directly produces a concise final solution, using 2977 tokens in total.}
\label{case:AIME_polar}
\end{figure*}

\begin{figure*}
\centering
\begin{tcolorbox}[size=title,opacityfill=0.1]
\noindent
\faUser\ \textbf{Question (GPQA)}: Answer the following multiple choice question. The last line of your response should be of the following format: 'Answer: \$LETTER' (without quotes) where LETTER is one of ABCD. Think step by step before answering. "Oh, I know you," the ribonucleoprotein particle says to the nascent chain as they meet. "Pause there for a minute. Let me show you in; you really need some sugar." "It seems somewhat rough. I guess this is goodbye; I need to be on my way", the chain replies. Where did they meet, and where is the chain heading? A) Golgi to the mitochondrion B) membrane to the nucleus C) cytosol to the extracellular space D) ribosome to the proteasome. \\

\faCogs\ \textbf{Thinking}:

\texttt{\textbf{<think>}} I need to figure out where the ribonucleoprotein particle and the nascent chain meet, and where the chain is heading. The options are: A) Golgi to the mitochondrion B) membrane to the nucleus C) ribosome to the proteasome D) cytosol to the extracellular space First, I recall that the ribonucleoprotein particle (RNP) is involved in transporting nucleotides into cells. Nucleotides are the building blocks of DNA and RNA. The nascent chain is probably a nascent protein, which is a protein that's being synthesized by the ribosome. \textbf{[...](omit)} So, option D. Perhaps in some contexts, "cytosol" is used for the cell membrane. I think I'll go with D. But let's see if there's a better fit. Option A: "Golgi to the mitochondrion" – if the chain is heading to the Golgi, but why would it be at "Golgi to mitochondrion". Perhaps for a protein that is going to the mitochondrion, but the question is about a general nascent chain needing nucleotides, not necessarily for a specific location. So, unlikely. Perhaps in this case, after meeting, it's heading to the Golgi, and the meeting is at the cytoplasm, but not specified. I think D is the most plausible. Perhaps B: "membrane to the nucleus" – for a protein that is being integrated, but again, not for nucleotide entry. I think D is the answer. So, I'll say D. But let's confirm. Final answer. \texttt{\textbf{</think>}} The ribonucleoprotein particle (RNP) shuttles nucleotides (Sugar) into cells, and the nascent chain (a protein being synthesized) is present in the cytoplasm. \textbf{[...](omit)} - B) Membrane to the nucleus: Incorrect, as this is typically for secreted proteins that integrate into the nucleus, but the chain needs nucleotides for assembly, not nuclear integration. - C) Ribosome to the proteasome: Incorrect, as the ribosome and proteasome are locations for different processes; the ribosome assembles the protein, while proteasomal degradation occurs later. - D) Cytosol to extracellular space: Correct, as the meeting occurs at the cell membrane (boundary of cytoplasm and extracellular space), and the chain heads to the extracellular space after receiving the nucleotides. Thus, they meet at the cell membrane (described as "cytosol to extracellular space"), and the chain is heading to the extracellular space. \textbf{(22149 tokens)}
\
\\

{\color[HTML]{2EA121}\faCogs\ \textbf{Ours}}: \\
\texttt{\textbf{/nothink <think></think>}} Let's think step by step to answer the question. Understanding the Question: The question is a multiple-choice question about the meeting place and destination of a nascent (newly synthesized) protein chain. The options are: - A) Golgi to the mitochondrion  \textbf{[...](omit)} This fits because the cytoplasm (where the ribonucleoprotein is located) processes the protein, and the processed protein then exits the cell to the extracellular space. - D) Ribosome to the proteasome: The ribosome is where protein synthesis occurs, and the proteasome is involved in protein degradation. The described process is not about moving from the ribosome to the proteasome but about processing and exiting. Conclusion: The correct answer is C) cytosol to the extracellular space, as the ribonucleoprotein processes the nascent protein in the cytoplasm, and the processed protein is then released into the extracellular space. \textbf{(2197 tokens)}

\end{tcolorbox}
\caption{For an example from GPQA, the Thinking baseline generates a long chain-of-thought with redundant intermediate steps. In contrast, our method chooses \textit{NoThinking} and directly produces a concise final solution, using 2197 tokens in total.}
\label{case:GPQA_polar}
\end{figure*}

\end{document}